\documentclass[letterpaper, 10 pt, conference]{ieeeconf}  

\IEEEoverridecommandlockouts                              

\usepackage{amsmath}
\usepackage{amssymb}  
\usepackage[final]{graphicx}
\usepackage{bm}
\usepackage{cite}
\usepackage[tracking]{microtype}
\usepackage{csquotes}
\usepackage[draft,inline,nomargin,index]{fixme}
\usepackage{mathtools}
\usepackage{hyperref}
\usepackage[super]{nth}
\usepackage[shortcuts, acronym, toc, nonumberlist=false]{glossaries}
\makenoidxglossaries

\newcommand{\etal}{et al.\ }
\newcommand{\ie}{i.\,e.,\ }

\newcommand{\eg}{e.\,g.,\ }

\usepackage{booktabs}
\usepackage{boldline}
\usepackage{siunitx}
\DeclareSIUnit{\rad}{rad}
\newacronym
 {rmse}{RMSE}{Root Mean Squared Error}
\newacronym
 {sl}{SL}{Supervised Learning}
\newacronym
 {mlp}{MLP}{multi-layer perceptron}
 \newacronym{hdmap}{HD Map}{High-Definition Map}
\newacronym{sft}{SFT}{Symmetric Fusion Transformer}
\newacronym{rpe}{RPE}{relative positional embedding}
\newacronym{wta}{WTA}{Winner-Takes-All}
\newacronym{cvae}{CVAE}{Conditional Variational Autoencoder}
\newacronym{elbo}{ELBO}{Evidence Lower Bound}
\newacronym{kl}{KL}{Kullback-Leibler}
\newacronym{ade}{ADE}{Average Displacement Error}
\newacronym{sade}{SADE}{Scene Average Displacement Error}
\newacronym{fde}{FDE}{Final Displacement Error}
\newacronym{sfde}{SFDE}{Scene Final Displacement Error}
\newacronym{amr}{actorMR}{Actor Miss Rate}
\newacronym{acr}{actorCR}{Actor Collision Rate}
\newacronym{lr}{LR}{Learning Rate}
\newacronym{dl}{DL}{Deep Learning}
\newacronym{mhca}{MHCA}{Multi-Head Cross Attention}
\newacronym{mhsa}{MHSA}{Multi-Head Self Attention}
\newacronym{ffn}{FFN}{Feed-Forward Network}

\title{\LARGE \bf
From Marginal to Joint Predictions: Evaluating Scene-Consistent Trajectory Prediction Approaches for Automated Driving
}

\author{Fabian Konstantinidis$^{1,3*}$,
        Ariel Dallari Guerreiro$^{2*}$,
        Raphael Trumpp$^{2}$,
        Moritz Sackmann$^{1}$,\\
        Ulrich Hofmann$^{1}$,
        Marco Caccamo$^{2}$,
        Christoph Stiller$^{3}$
\thanks{This work is a result of the joint research project STADT:up (19A22006E). The project is supported by the German Federal Ministry for Economic Affairs and Climate Action (BMWK), based on a decision of the German Bundestag. The author is solely responsible for the content of this publication.}
\thanks{Marco Caccamo was supported by an Alexander von Humboldt Professorship endowed by the German Federal Ministry of Education and Research.}
\thanks{$^{1}$Pre-Development of Automated Driving, CARIAD SE, 38442 Wolfsburg, Germany. Email: {\tt\small \{fabian.konstantinidis, moritz.sackmann, ulrich.hofmann\}.@cariad.technology}}%
\thanks{$^{2}$Technical University of Munich (TUM), 85748 Garching, Germany. Email: {\tt\small \{ariel.guerreiro, raphael.trumpp, mcaccamo\}@tum.de}}%
\thanks{$^{3}$Karlsruhe Institute of Technology (KIT), 76131 Karlsruhe, Germany. Email: {\tt\small christoph.stiller@kit.edu}}%
\thanks{* Equal contribution}%
}

\begin{document}
\maketitle
\thispagestyle{empty}
\pagestyle{empty}

\begin{abstract}
Accurate motion prediction of surrounding traffic participants is crucial for the safe and efficient operation of automated vehicles in dynamic environments. Marginal prediction models commonly forecast each agent's future trajectories independently, often leading to sub-optimal planning decisions for an automated vehicle. In contrast, joint prediction models explicitly account for the interactions between agents, yielding socially and physically consistent predictions on a scene level. However, existing approaches differ not only in their problem formulation but also in the model architectures and implementation details used, making it difficult to compare them. In this work, we systematically investigate different approaches to joint motion prediction, including post-processing of the marginal predictions, explicitly training the model for joint predictions, and framing the problem as a generative task. We evaluate each approach in terms of prediction accuracy, multi-modality, and inference efficiency, offering a comprehensive analysis of the strengths and limitations of each approach. Several prediction examples are available at \href{https://frommarginaltojointpred.github.io/}{https://frommarginaltojointpred.github.io/}.
\end{abstract}

\section{INTRODUCTION}

Automated vehicles operate in highly dynamic environments, making the anticipation of the future behavior of surrounding traffic participants crucial for safe and efficient decision-making.

\begin{figure}[t]
  \centering
  \includegraphics[width=\linewidth]{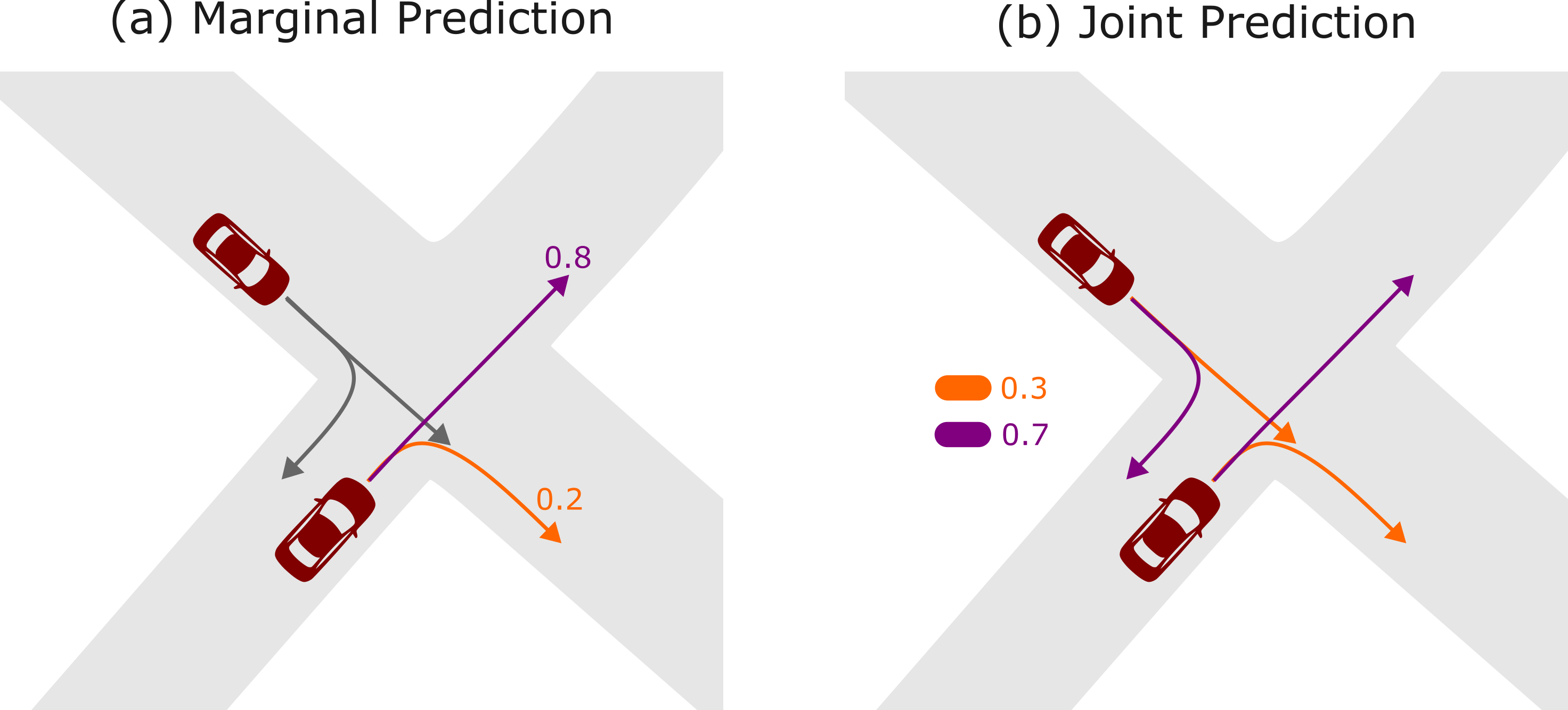}
  \caption{Comparison of marginal predictions and joint predictions for automated driving. Marginal predictions (\textbf{left}) generate $K$ future predictions independently per agent, with each prediction being scored per agent. Conversely, joint predictions (\textbf{right}) model all agents’ trajectories together, capturing their interactions and dependencies. We use $K=2$ for clarity.}
  \label{fig:example_image}
\end{figure}

Existing approaches to motion prediction commonly tackle this challenge by forecasting each agent's future trajectories independently, a strategy known as \emph{marginal prediction} \cite{zhang2024simpl, vectornet, densetnt}. While these methods are effective in predicting the distribution over future trajectories for individual agents, these methods often fail to capture the complex interactions between road users. This can lead to predicted trajectories that may be accurate individually, but implausible when considered jointly, \eg two predicted vehicles crossing an intersection at the same time, as depicted in Figure \ref{fig:example_image}a, which would result in a collision. Due to these issues, an automated vehicle might respond suboptimally, \eg by braking unnecessarily strongly, adopt a very conservative driving strategy, or even worse, execute unsafe plans based on faulty assumptions \cite{hagedorn2024integration}.

In contrast, \emph{joint prediction} models explicitly capture the dependencies between agents, enabling the generation of socially and physically plausible multi-agent futures, as depicted in Figure \ref{fig:example_image}b. By reasoning over the collective behavior of all agents, these models produce predictions that not only adhere to scene constraints, \eg respecting right-of-way rules in multi-agent merges, but also reflect the kinds of strategic interactions that are common in complex traffic scenarios. Thus, joint predictions offer more reliable inputs for downstream planning, reduce the risk of infeasible or conflicting behaviors, and better align with the real-world requirements of automated driving systems. Additionally, joint models can learn correlated uncertainties, capturing how changes in one agent's behavior influence others.

Moreover, existing approaches to joint motion prediction differ not only in their problem formulation, such as recombining marginal predictions into joint predictions \cite{shi2024mtr++, sun2022m2i, gilles2021thomas}, using scene-level losses during training \cite{Ngiam2022SceneTransformer, wagner2024scenemotion}, or adopting generative formulations \cite{motionlm, casas2020implicit, lookout}, but also in their model architectures, dataset pre-processing steps, and post-processing techniques. As a result, it is difficult to determine whether gains in leaderboard performance result from different problem formulations, architectural changes, or other implementation details such as training strategies. We address this ambiguity by providing a detailed evaluation of commonly used approaches to joint motion prediction.

\emph{\textbf{Our contributions}} are four-fold: (i) We systematically explore \emph{joint} prediction strategies building upon the SIMPL model \cite{zhang2024simpl} as a marginal baseline. (ii) We outline several possible modifications to adapt SIMPL for joint prediction, including the use of more expressive trajectory decoders, as well as framing joint motion prediction as a generative task using a \gls{cvae}. (iii) We conduct extensive experiments on the Argoverse 2 Motion Forecasting Dataset \cite{Argoverse2}, evaluating not only prediction accuracy but also the multi-modality of the predicted modes and the models' inference times. (iv) By highlighting the strengths and limitations of each approach, we provide a fair comparison between the approaches.

\section{RELATED WORK}

Marginal prediction models, which forecast each agent's future multi-modal trajectories independently, have already been extensively studied, \eg \cite{vectornet, zhang2024simpl, shi2024mtr++, densetnt, wayformer}.
A prominent example is Vectornet \cite{vectornet}, which introduces a vectorized scene representation, where agents and map elements are modeled as sets of vectors. These are processed through a hierarchical graph neural network to predict individual trajectories. Due to this input representation, Vectornet can be flexibly applied to a diverse set of traffic scenarios. Many follow-up approaches build upon VectorNet, adopting a similar input representation, \eg \cite{densetnt, shi2024mtr++, zhang2024simpl}. To further enhance the marginal prediction performance, many recent works leverage Transformer-based architectures, \eg \cite{zhang2024simpl, zhang2023hptr, wayformer, philion2024trajeglish, zhou2023query}. Notably, SIMPL \cite{zhang2024simpl} and MTR++ \cite{shi2024mtr++} do not process the scene from every predicted agent's point of view to generate its corresponding marginal predictions. Instead, they adopt an instance-centric scene representation, where each instance, \ie an agent or map element, is represented in its own local coordinate frame. Utilizing a relative positional encoding that captures the relative positions between instances, they enable an efficient and viewpoint-invariant processing of the full scene. 

Joint prediction models explicitly model the distribution over future motions for multiple agents, \eg \cite{wagner2024scenemotion, jia2023hdgt, liu2024laformer, Liu2021StackedTransformer, Ngiam2022SceneTransformer, janjovs2022san, casas2020implicit, lookout}. A naive approach to joint predictions is the \emph{recombination of marginally predicted modes} into joint predictions, as done in \cite{shi2024mtr++, gilles2021thomas, sun2022m2i}. In \cite{shi2024mtr++} and \cite{sun2022m2i}, the top-$K$ combinations of marginal predictions are selected based on the product of their confidence scores. Gilles \etal \cite{gilles2021thomas} propose a more sophisticated approach, where the recombination of marginal modes is realized with a learnable recombination module that produces consistent, reordered combinations.

As originally proposed in \cite{Ngiam2022SceneTransformer}, another approach to joint motion prediction is to train the model using a \emph{scene-level loss} to directly output groups of trajectories, evaluated as joint trajectory modes. This idea is adopted in many works nowadays, \eg \cite{wagner2024scenemotion, jia2023hdgt, liu2024laformer, Liu2021StackedTransformer}. Ngiam \etal \cite{Ngiam2022SceneTransformer} propose to aggregate the per-agent loss across all agents of a predicted mode, resulting in a loss formulation on the scene-level.

Alternatively, the task of motion prediction can be framed as a \emph{generative task}, where a generative model is learned to capture the full distribution of possible future outcomes, \eg \cite{motionlm, casas2020implicit, lookout}. For instance, Seff \etal \cite{motionlm} propose to frame multi-agent motion forecasting as a language modeling task by autoregressively predicting distributions over discrete motion tokens and sampling from them in every forecasting step. An alternative formulation is explored in \cite{casas2020implicit} and its follow-up work \cite{lookout}, where the problem of joint motion prediction is solved via a \gls{cvae}-based model. Encoding the scene into distributions over latent scene representations enables sampling from them. Combined with a deterministic decoder processing the drawn samples, multi-agent trajectories are produced that are consistent across traffic participants.

In this work, due to its efficiency, we adopt SIMPL \cite{zhang2024simpl} as our marginal baseline model and extend it systematically to joint predictions by extending it with the discussed approaches, thus enabling a fair comparison of these approaches.

\section{BACKGROUND}

The task of trajectory prediction involves forecasting the future trajectories $\mathbf{Y} = \{\mathbf{y_1}, \mathbf{y_2}, \dots, \mathbf{y_{N_a}}\}$ of $N_a$ traffic participants given their past trajectories $\mathbf{X} = \{\mathbf{x_1}, \mathbf{x_2}, \dots, \mathbf{x_{N_a}}\}$ and static map information $\mathcal{M}$. Each agent’s past and future trajectories are given by $\mathbf{x}_i = \{x_{i,t} \mid t = -T_p + 1, \dots, 0\}$ and $\mathbf{y}_i = \{y_{i,t} \mid t = 1, \dots, T\}$, representing $T_p$ historical and $T$ future time steps, respectively. Commonly, in addition to the agent's position, additional relevant features are included in $\mathbf{x}_i$ , such as its heading, velocity, and type. The static map information $\mathcal{M}$ is typically given as a \gls{hdmap}. 

To generate predictions of future trajectories, a model must find the optimal parameters $\theta^*$ such that the learned distribution over trajectories $P_{\theta^*}$ closely resembles the true distribution of trajectories. Formally, this can be expressed as
\begin{equation}
    \theta^* = \arg\min_\theta \mathbb{E}_{\mathbf{\hat{Y} \sim P_\theta \left(\cdot | \mathbf{X}, \mathcal{M} \right)}} \left[ \mathcal{L}(\mathbf{Y}, \mathbf{\hat{Y}}) \right],
\end{equation}
where $\mathcal{L}$ quantifies the discrepancy between the predicted trajectories $\mathbf{\hat{Y}}$ drawn from $P_\theta$ and the corresponding ground truth trajectories $\mathbf{Y}$.

For multi-modal, joint prediction, the model must be able to generate multiple scene-level trajectories $\mathbf{\hat{Y}}$, \ie
\begin{equation}
    \label{eq2:multimodal_joint}
    P_\theta(\mathbf{\hat{Y}}|\mathbf{X},\mathcal{M}) = \sum_{k = 1}^\mathbf{K} \alpha^k P_\theta(\mathbf{\hat{Y}^k}|\mathbf{X},\mathcal{M}),
\end{equation}
where $K$ denotes the number of predicted scene-level modes and $\alpha^k$ is the confidence score of the $k$-th mode, considered as an auxiliary task to trajectory forecasting.

\subsection{Marginal Baseline Model} 

In this work, we adopt SIMPL \cite{zhang2024simpl} as our baseline model. SIMPL is an efficient, open-source framework for multi-agent motion prediction, focusing on marginal predictions. We briefly introduce this framework below and refer to \cite{zhang2024simpl} for a more detailed explanation.

SIMPL utilizes an instance-centric scene representation, where the features of an instance, \eg an actor or map element, are expressed relative to the instance's local reference frame. For actors, the frame is centered at its last observed position with its heading aligned to the x-axis. Map elements are modeled as polylines, using their centroid as the origin and aligning the frame with the displacement vector between its endpoints. Actor features are encoded via a 1D CNN-based Actor Encoder \cite{LaneNet}, producing latent actor tokens $Z^\mathrm{(actor)} = \{z_{i}^{\mathrm{(actor)}} \mid i = 1, \dots, N_a\}$. Similarly, map elements are encoded into the same latent space by a PointNet-based Map Encoder \cite{vectornet, qi2017pointnet}, generating map tokens $Z^\mathrm{(map)} = \{z_{i}^{\mathrm{(map)}} \mid i = 1, \dots, N_m\}$, where $N_m$ denotes the number of polylines in $\mathcal{M}$. 

To model spatial relations, the pairwise relative position $r_{i \rightarrow j}$ between instances $i$ and $j$ is described using the heading difference $\alpha_{i \rightarrow j}$, relative azimuth $\beta_{i \rightarrow j}$, and Euclidean distance $\left\| \mathbf{p}_{i \rightarrow j} \right\|$ between the origins of the reference frames. These are embedded into an \gls{rpe} $r_{i \rightarrow j}' \in \mathbb{R}^{D}$ by passing $r_{i \rightarrow j}$ through an MLP.

Interaction among tokens is modeled using multiple layers of an \gls{sft} \cite{zhang2024simpl}, applied to the combined token set $[Z^\mathrm{(actor)}, Z^\mathrm{(map)}]$. Each \gls{sft} layer includes a \gls{mhca} with a skip connection, followed by layer normalization, a \gls{ffn} with another skip connection, and a second layer normalization. In the \gls{mhca}, the tokens in $Z$ serve as queries, while the keys and values are derived from the relative context tokens $c_{i \rightarrow j} = \mathrm{MLP}(z_i \oplus z_j \oplus, r_{i \rightarrow j}')$, where $\oplus$ denotes a concatenation. Each relative embedding $r_{i \rightarrow j}'$ is further refined by re-encoding its associated context token $c_{i \rightarrow j}$ with another MLP and adding the result back to $r_{i \rightarrow j}'$. A single \gls{sft}-layer is expressed as:
\begin{equation}\label{eq:sft}
    \{Z^\mathrm{(k+1)}, \mathrm{M}_{rpe}^\mathrm{(k+1)} \} = \mathrm{SFT}\left(Z^\mathrm{(k)},\mathrm{M}_{rpe}^\mathrm{(k)}\right),
\end{equation}
where $\mathrm{M}_{rpe}^\mathrm{(k)} \in \mathbb{R}^{N \times N \times D}$  is the matrix of \gls{rpe}s $r_{i \rightarrow j}'$ used in the $k$-th \gls{sft} layer, with $N$ denoting the number of tokens in $Z$ and $D$ the hidden dimension. The first \gls{sft} layer is initialized with $Z^{(1)}=[Z^\mathrm{(actor)}, Z^\mathrm{(map)}].$

Lastly, the refined actor tokens $Z^{\mathrm{(actor)}}$ are then split into the $K$ embeddings, each passed through a trajectory-decoder, realized as an \gls{mlp}, to predict one possible future trajectory as well as the corresponding confidence score $\alpha^k$. However, instead of directly predicting the $T$ future waypoints, the control points for a Bézier curve are predicted.

The model is trained using a \gls{wta} loss, where for each agent, only the best-matching predicted trajectory is used to compute the optimization loss.

\section{EVALUATED APPROACHES} \label{sec:method}
Our objective is to identify the most effective approach to joint motion prediction. To this end, we explore three strategies: (i) recombining marginal modes into joint predictions, (ii) applying a scene-level loss during training, and (iii) formulating motion prediction as a generative problem solved via a \gls{cvae}. 

Specifically, we evaluate these approaches by comparing the following models:
\begin{enumerate} 
    \item \emph{Marginal Recombination} (Section \ref{sec:approach_recombination}): The marginal baseline model evaluated jointly by recombining predicted marginal modes. 
    \item \emph{Joint Loss} (Section \ref{sec:method_jointloss}): The baseline model trained directly with a scene-level loss to encourage joint consistency. 
    \item \emph{Multi-MLP} (Section \ref{sec:method_jointloss}): An extension of the baseline with a Multi-MLP decoder, using separate heads for each mode, trained with the scene-level loss. 
    \item \emph{Anchor Point Transformer} (Section \ref{sec:method_jointloss}): An extension of the baseline with a Transformer-based decoder to improve inter-mode accuracy, also trained with the scene-level loss. 
    \item \emph{CVAE} (Section \ref{sec:approach_generative}): The baseline model adapted to the \gls{cvae} framework, enabling sampling-based generation of predictions. 
\end{enumerate}

\subsection{Recombination of Marginal Modes}\label{sec:approach_recombination}
Our first approach for extending the baseline model to joint predictions does not involve any retraining. Instead, following the method proposed in \cite{sun2022m2i, shi2024mtr++}, we simply recombine the marginal predictions produced by the baseline model into joint predictions. Specifically, for $N_a$ agents, each predicting $K$ marginal modes, we construct all $K^{N_a}$ possible joint combinations. We then select the top $K$ joint modes based on a joint confidence score $\alpha_{\text{scene}}^{\mathbf{k}}$, defined as: 
\begin{equation} 
    \alpha_{\text{scene}}^{\mathbf{k}} = \prod_{i=1}^{N_a} \alpha_i^{k_i}, 
\end{equation} 
where $\mathbf{k} = \{k_1, \dots, k_{N_a}\}$ indexes the selected marginal mode for each agent, and $\alpha_i^{k_i}$ is the confidence score of agent $i$'s $k_i$-th marginal mode.

While this simple technique does not require any retraining of the model, it necessitates computationally expensive post-processing, which increases with the number of actors in the scene. To minimize computational overhead, we implemented this as a beam search, calculating the top-$K$ scene confidence scores stepwise by expanding only the most probable joint combinations in each step. 

\begin{figure*}[!t]
    \vspace{4pt}
    \centering
    \includegraphics[width=0.963\textwidth]{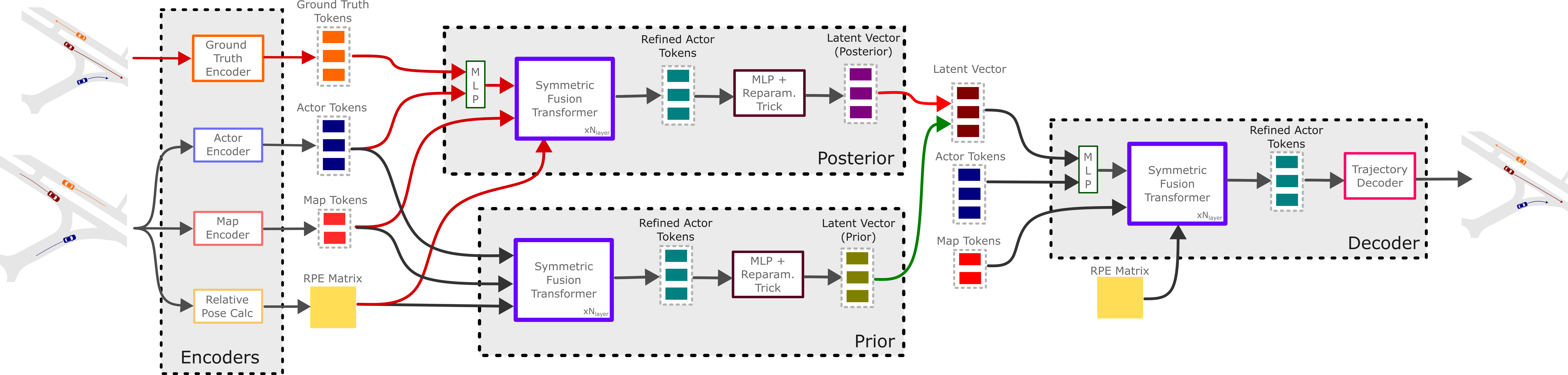}
    \caption{Illustration of the proposed \gls{cvae} architecture, composed of the \emph{Encoder}, \emph{Posterior}, \emph{Prior}, and \emph{Decoder} submodules. Here, the unused outputs generated by the \gls{sft} are suppressed. The paths indicated with black arrows are always executed, while the paths with red arrows are only executed during training, and the green path is only executed during inference.}
    \label{fig:cvae_model}
    \vspace{-3pt}
\end{figure*}

\subsection{Scene-Level Loss}\label{sec:method_jointloss}

As an alternative that does not require any computationally expensive post-processing, commonly, the training objective is modified to calculate the loss on a scene-level instead of an actor-level. An actor-level loss treats each actor independently using the mode best aligned with the ground truth for training, thus focusing only on per-agent accuracy. However, when predicting trajectories jointly, coordination and consistency between the trajectories of a mode are required. As proposed in \cite{Ngiam2022SceneTransformer}, we achieve this by grouping the predicted modes of the agents into $K$ scenes, where every agent's $k$-th predicted trajectory contributes to the $k$-th scene, and calculating a loss over all agents in this mode. 

Specifically, we adopt the loss formulation proposed in \cite{jia2023hdgt}, which combines a regression loss $\mathcal{L}_{reg}$ and a classification loss $\mathcal{L}_{cls}$, weighted by $\lambda_{reg}$ and $\lambda_{cls}$, respectively. The regression loss is defined as:
\begin{equation}\label{eq:joint_loss_reg}
    \mathcal{L}_{reg} = \min_k\frac{1}{N_a\cdot T} \sum_{n = 1}^{N_a} \sum_{t=1}^{T}  \ell(\hat{\mathbf{y}}_{n, t}^k, \mathbf{y}_{n, t} ),
\end{equation}
where $\ell(\cdot)$ denotes the Smooth L1 loss \cite{Ross2025SmoothL1}. This loss is averaged over the $T$ predicted timesteps and the $N_a$ agents for the $k$-th predicted scene mode. The $\min$ operator implements the \gls{wta} strategy at the scene level, selecting the prediction that best matches the ground-truth joint future.

The classification loss, used for confidence scoring, is defined as the Cross-Entropy loss between the scene-level confidence scores $\alpha \in \mathbb{R}^K$ and the one-hot encoding of the mode $k^*$ that minimizes the regression loss.

When extending the marginal model to joint predictions by using a scene-level loss, the decoder not only has to reason about where a single agent wants to drive, but about the futures of multiple actors jointly, requiring a deeper understanding of interactions and cooperation between agents. Thus, we propose two architectural modifications to the decoder used in \cite{zhang2024simpl}, to better account for the joint multi-modal structure of the problem.

\textbf{\emph{Multi-\gls{mlp} Decoder}}:
When using a single decoder-\gls{mlp}, even when trained with the \gls{wta}-loss, it would be forced to update its weights to improve the performance on all predictions, resulting in some form of weight-averaging. To mitigate this issue, we follow the recommendation in \cite{janjovs2022san} and increase the model's capacity to reduce the interference between predictions during training. This is done by using a separate decoder-\gls{mlp} for every scene mode being predicted, resulting in a clear separation between the decoder-weights responsible for predicting a mode.

\textbf{\emph{Anchor Point Transformer Decoder}}: 
The Multi-\gls{mlp} decoder effectively addresses the issue of weight-averaging, but allows no coordination between the modes. To further enhance the trajectory decoder, we improve its inter-mode awareness by explicitly modeling the dependencies between modes, similarly to \cite{wagner2024scenemotion, zhang2023hptr}. This is done by introducing a set of learnable anchors, $A^{(1)} \in \mathbb{R}^{K\times D}$, which are refined for each agent through multiple Transformer-decoder layers \cite{vaswani2017attention}. Each Transformer-decoder layer consists of a \gls{mhsa} mechanism to model mode dependencies, followed by \gls{mhca} with the refined actor tokens $Z^{\mathrm{(actor)}}$ as keys and values, and a \gls{ffn}. Each component is implemented with a skip connection and layer normalization. After $L$ layers, the refined anchors $A^{(L)} \in \mathbb{R}^{N_a\times K\times D}$ are passed into the Multi-MLP decoder to produce the final predictions.

\subsection{Generative Problem Formulation}\label{sec:approach_generative} 

While the approaches proposed in Section \ref{sec:method_jointloss} are explicitly trained to generate accurate joint predictions, they are supervised using the \gls{wta}-loss in Equation (\ref{eq:joint_loss_reg}), where optimization is only applied to the mode closest to the ground truth. Consequently, the remaining predictions receive no gradient feedback, potentially resulting in implausible predictions and inefficient use of model capacity unless additional regularization, \eg diversity losses \cite{varadarajan2022multipath++}, is introduced. 

To address these challenges, prior works, \eg \cite{casas2020implicit, motionlm, lookout, choi2022hierarchical}, reframe motion prediction as a generative task, training models to capture the full distribution of possible future outcomes. While this requires explicit modifications to the baseline model, it offers key advantages: instead of predicting a fixed number $K$ of modes, the model generates diverse, plausible samples, each representing one predicted scene mode. Moreover, the number of samples drawn can be arbitrarily scaled and dynamically adjusted to the available computation time, making it well-suited for real-time applications.

Specifically, we investigate formulating the motion prediction task as a \gls{cvae} approach, as proposed in \cite{casas2020implicit}. Our proposed \gls{cvae}-based architecture, as depicted in Figure \ref{fig:cvae_model}, is organized into four key modules: \emph{Instance-Encoders}, \emph{Posterior Network}, \emph{Prior Network}, and a \emph{Deterministic Decoder}. 

The instance encoders encode the actors' past trajectories $\mathbf{X}$, the static map information $\mathcal{M}$, and the actors' ground truth future trajectories $\mathbf{Y}$ into a set of latent tokens serving as input for the subsequent modules.

The encoded instance tokens are then used for inferring the posterior $q(\mathbf{B}\mid\mathbf{Y}, \mathbf{X}, \mathcal{M})$ and prior $p(\mathbf{B}\mid\mathbf{X}, \mathcal{M})$ distributions over the latent scene variables $\mathbf{B}$, which capture the uncertainty of the joint scene prediction. Specifically, $\mathbf{B}=\{\mathbf{b}_i\in\mathbb{R}^{D_B}\mid i=1,\cdots,N_a\}$ represents a set of actor-specific latent tokens used to condition the deterministic decoder. Intuitively, each latent token $\mathbf{b}$ represents the multi-modal future of an actor, including its high-level driving goals and individual driving style. 

As proposed in \cite{casas2020implicit}, conditioning the deterministic decoder with the latent scene variables $\mathbf{B}$ produces an implicit joint prediction of the scene $\hat{\mathbf{Y}}=f(\mathbf{X}, \mathcal{M}, \mathbf{B})$. 
During training, $\mathbf{B}$ is sampled from the posterior distribution, while during inference, it is sampled from the prior.

Unlike the approaches presented in Section \ref{sec:method_jointloss}, which directly generate a fixed number $K$ of trajectory candidates, our \gls{cvae} decoder produces only a single mode per sampled latent $\mathbf{B}$. Thus, the multi-modality of our \gls{cvae} model stems not from the decoder itself, but from sampling multiple latent representations $\{\mathbf{B}_k \mid 1, \cdots, K\}$ from the prior. Consequently, the prediction uncertainty is captured explicitly through the latent distribution, decoupling it from the deterministic decoder.

\textbf{\emph{Modified Model}}: 
As in the baseline model \cite{zhang2024simpl}, we use a CNN-based Actor Encoder \cite{LaneNet} to generate the actor tokens $Z^{(\mathrm{actor})}$, a PointNet-based Map Encoder \cite{qi2017pointnet} for generating the map tokens $Z^{(\mathrm{map})}$, and the \gls{rpe} matrix $\mathrm{M}_{rpe}$ is obtained via a simple \gls{mlp}. In addition, the actors' future trajectories are encoded into $Z^{(\mathrm{actor, gt})}$ using an additional CNN-based Actor Encoder \cite{LaneNet}.

Following \cite{casas2020implicit}, we define the posterior distribution as a multivariate Gaussian with a diagonal covariance matrix for each agent. The posterior network $q_\phi(\mathbf{B}\mid \mathbf{Y}, \mathbf{X}, \mathcal{M})$ approximates this distribution by processing the combined tokens $[Z^{(\mathrm{actor, gt})}, Z^{(\mathrm{actor})}, Z^{(\mathrm{map})}]$ to produce the mean $\mu_\mathrm{post} \in \mathbb{R}^{N_a \times D_B}$ and the entries of a diagonal covariance matrix $\sigma^2_\mathrm{post} \in \mathbb{R}^{N_a \times D_B}$. 

The posterior network consists of multiple layers of \gls{sft}, expressed by (\ref{eq:sft}), followed by an \gls{mlp} that predicts the distribution parameters from the refined actor tokens. The first \gls{sft} layer is initialized with $Z^{(1)}_\mathrm{post} = [\hat{Z}^\mathrm{(actor)}, Z^\mathrm{(map)}]$, where the augmented actor tokens $\hat{Z}^\mathrm{(actor)} = \mathrm{MLP}_\phi\left(Z^{(\mathrm{actor, gt})} \oplus Z^{(\mathrm{actor})} \right)$ are obtained by projecting the concatenated actor tokens, capturing the past and future trajectories of each agent, back to the original dimension of  $Z^{(\mathrm{actor})}$. After several \gls{sft} layers, each refined actor token is individually processed by another \gls{mlp} to generate its corresponding row in $\mu_\mathrm{post}$ and $\sigma^2_{\mathrm{post}}$.

Although each latent token $\mathbf{b}_i \sim \mathcal{N}\left( \mu_\mathrm{post, i}, \mathrm{diag}(\sigma^2_\mathrm{post, i})\right)$ is anchored to a specific agent, it encodes information about the entire scene. This is because, through the message propagation in the \gls{sft} layers, each refined actor token depends on the full input $[Z^{(\mathrm{actor, gt})}, Z^{(\mathrm{actor})}, Z^{(\mathrm{map})}]$, enabling holistic scene understanding at the token level.

The prior network $p_\psi(\mathbf{B}\mid \mathbf{X}, \mathcal{M})$ has the same structure as the posterior network, with the only difference being the initialization of the first \gls{sft} layer. Since in the prior we do not have access to the ground-truth tokens $Z^{(\mathrm{actor, gt})}$, we set $Z^{(1)}_\mathrm{prior} = [Z^\mathrm{(actor)}, Z^\mathrm{(map)}]$.

Similarly, the deterministic decoder $f_\omega(\mathbf{X}, \mathcal{M}, \mathbf{B})$ is also realized through multiple layers of \gls{sft}, followed by a trajectory decoder that transforms the refined actor tokens into one possible future prediction $\hat{\mathbf{Y}}$. Its first \gls{sft} layer is initialized with $Z^{(1)}_\mathrm{dec} = [\hat{Z}^\mathrm{(actor)}, Z^\mathrm{(map)}]$, where the augmented actor tokens $\hat{Z}^\mathrm{(actor)} = \mathrm{MLP}_\omega\left( Z^{(\mathrm{actor})} \oplus \mathbf{B} \right)$ are obtained by projecting the concatenation of the actor tokens and their corresponding entries in the scene latent $\mathbf{B}$ back into the original dimension of  $Z^{(\mathrm{actor})}$. Since the deterministic decoder predicts only a single realization of the scene, we retain the trajectory decoder from the baseline \cite{zhang2024simpl}.

\textbf{\emph{Model Training}}:
The trainable parameters of the prior, posterior, and decoder networks are denoted as $\psi$, $\phi$, and $\omega$, respectively. Following \cite{casas2020implicit}, we optimize the model's parameters by minimizing the modified \gls{elbo} objective:
\begin{equation}\label{eq:loss_elbo} 
    \mathcal{L}_{\mathrm{ELBO}} = \mathcal{L}_{\mathrm{reg}} + \beta \cdot \mathcal{L}_{\mathrm{KL}}, 
\end{equation}
where $\mathcal{L}_{\mathrm{reg}}$ is the regression loss from Equation (\ref{eq:joint_loss_reg}) with $K=1$, and $\mathcal{L}_{\mathrm{KL}}$ is the Kullback Leiber divergence between $p_\psi(\mathbf{B}\mid \mathbf{X}, \mathcal{M})$ and $q_\phi(\mathbf{B}\mid \mathbf{Y},\mathbf{X}, \mathcal{M})$. The hyperparameter $\beta$ controls the trade-off between both losses.

\begin{table*}[t]
    \vspace{4pt}
    \centering
    \caption{Prediction Performance on the Argoverse 2 Motion Forecasting Competition Test Set.}
    \begin{tabular}{ccccc}
        \toprule
        \textbf{Model} & \textbf{\begin{tabular}[c]{@{}c@{}}minSFDE\\ (K=6)\end{tabular}} & \textbf{\begin{tabular}[c]{@{}c@{}}minSADE\\ (K=6)\end{tabular}} & \textbf{\begin{tabular}[c]{@{}c@{}}actorMR\\ (K=6)\end{tabular}} & \textbf{\begin{tabular}[c]{@{}c@{}}actorCR\\ (K=6)\end{tabular}} \\
        \midrule
        (1) Marginal Recombination & $2.460$ & $0.978$ & $0.239$ & $0.0094$ \\
        (2) Joint Loss & $2.515$ & $1.026$ & $0.266$ & $0.0093$ \\
        (3) Multi-MLP & $2.322$ & $0.947$ & $0.256$ & $0.0091$ \\
        (4) Anchor Point Transformer & $2.197$ & $0.904$ & $0.245$ & $\mathbf{0.0085}$ \\
        (5a) CVAE Small Beta\textsuperscript{\textdagger} & $2.064_{\pm 0.006}$ & $0.860_{\pm 0.002}$ & $0.252_{\pm 0.000}$ & $0.0165_{\pm 0.0001}$ \\
        \(\text{{(5b) CVAE Small Beta}}^{\dagger,\ddagger}\) & $\mathbf{2.014_{\pm 0.008}}$ & $\mathbf{0.844_{\pm 0.003}}$ & $0.241_{\pm 0.000}$ & $0.0153_{\pm 0.0003}$ \\
        (6a) CVAE Large Beta\textsuperscript{\textdagger} & $2.092_{\pm 0.001}$ & $0.868_{\pm 0.001}$ & $0.241_{\pm 0.000}$ & $0.0121_{\pm 0.0001}$ \\
        (6b) \(\text{CVAE Large Beta}^{\dagger,\ddagger}\) & $2.112_{\pm 0.004}$ & $0.875_{\pm 0.001}$ & $\mathbf{0.238_{\pm 0.001}}$ & $0.0114_{\pm 0.0001}$ \\
        \bottomrule
    \end{tabular}
    \vspace{-3pt}
    {\parbox{0.693\linewidth}{
        \vspace{1pt}
        \scriptsize
        $\dagger$ denotes generative models. Metrics are reported as mean and standard deviation across three submissions. \newline$\ddagger$ denotes that the prior mean is used for generating one mode.
    }}
    \label{tab5:test_metrics}
\end{table*}

\section{EXPERIMENTS}

\textbf{\emph{Dataset}}:
We use the \emph{Argoverse 2 Motion Forecasting Dataset} \cite{Argoverse2} for our experiments, which comprises \num{250000} scenarios, each with a duration of \SI{11}{s} sampled at \SI{10}{Hz}. From each scenario, the first \SI{5}{s} are given as the past trajectories, while the forecasting horizon is \SI{6}{s}. We keep the pre-processing from our baseline model \cite{zhang2024simpl}.

\textbf{\emph{Evaluation Metrics}}:
For the evaluation of the discussed models, we use the metrics from the corresponding \emph{Argoverse 2 Motion Forecasting Competition} \cite{Argoverse2}. Specifically, we focus on its multi-world prediction metrics. These metrics include the \gls{sade} and \gls{sfde}, where the minimum over $K=6$ modes is reported. The competition also presents auxiliary metrics evaluating the intra-mode consistency, such as \gls{acr} and \gls{amr}, assessing prediction accuracy at an actor level.

\textbf{\emph{Evaluated Models}}:
We evaluate the different approaches to joint motion prediction by comparing the models defined in Section \ref{sec:method}. For the \gls{cvae} model, we use a setting of $\beta=0.05$ in Equation (\ref{eq:loss_elbo}), and additionally evaluate a second variant with a larger $\beta=0.5$, referred to as \emph{CVAE Small Beta} and \emph{CVAE Large Beta}, respectively.

We use a hidden dimension of $D=128$ for all models. Models (1)–(4) employ four \gls{sft} layers each. For the two \gls{cvae}-based models, the prior network, posterior network, and deterministic decoder each contain two \gls{sft} layers. During inference, only the prior network and deterministic decoder are active, resulting in four \gls{sft} layers across all evaluated models for consistency.

Initial ablation studies on a \SI{20}{\%} subset of the dataset determined two further design choices: for the Anchor Point Transformer (model (4)), we use two layers (evaluated against one and three layers), and for the \gls{cvae}-based models, we set the latent scene variable dimension to $D_B=32$ (evaluated against $D_B=64$).

\textbf{\emph{Training Details}}:
We train all models with a batch size of $16$ on a single NVIDIA Quadro RTX 6000 \SI{24}{GB} GPU. Optimization is performed using Adam, with an initial learning rate of $0.001$ that is gradually reduced to $0.0001$ after 40 epochs. Models (1)–(4) are trained for $50$ epochs. Since the two \gls{cvae}-based models had not yet converged after $50$ epochs, as indicated by their validation loss, we extended their training by an additional $30$ epochs to ensure fair comparison.

\subsection{Model Performance}
We first evaluate the models' performance quantitatively on the test set of the \emph{Argoverse 2 Motion Forecasting Competition} \cite{Argoverse2}. The results for each model are summarized in Table \ref{tab5:test_metrics}. For the two \gls{cvae}-based models, we submitted two variants each: (a) all $K=6$ modes are generated by conditioning the deterministic decoder on samples drawn from the prior distribution, and (b) the first mode is generated by conditioning on the prior mean, \ie setting $\mathbf{B}_1 = \mu_\mathrm{prior}$, while the remaining modes $\{\mathbf{B}_k \mid 2, \cdots, K\}$ are sampled from the prior distribution.

The baseline model with recombined marginal modes already achieves strong performance across all metrics.
However, surprisingly, applying a scene-level loss to explicitly encourage joint consistency in the predicted modes only reduced the collision rate slightly, while all other metrics deteriorated. This highlights that architectural adaptations are necessary, and an adaptation of the loss alone is not sufficient.

Our first modification, \ie employing a Multi-MLP trajectory decoder with a separate head per mode, led to consistent improvements across all metrics. 
Further and more significant gains were achieved by adding the Anchor Point Transformer before the Multi-MLP decoder to explicitly model the inter-mode dependencies.
Notably, this model achieves the lowest \gls{acr} among all models, demonstrating the benefit of message propagation between modes. 

All CVAE-based models outperform all deterministic models in terms of the positional errors minSFDE and minSADE.
The best results are achieved by the CVAE Small Beta model with mean sampling (Model 5b), which reduces the minSFDE and minSADE by \SI{18.1}{\%} and \SI{13.7}{\%}, respectively, compared to the jointly evaluated baseline model (Marginal Recombination). However, compared to the CVAE model trained with a larger $\beta=0.5$, (5b) exhibits slightly higher \gls{amr} and \gls{acr}. This trade-off can be explained by the training dynamics: using a smaller value for $\beta$ assigns a smaller weight for the KL loss in Equation (\ref{eq:loss_elbo}) and thus allows the prior to diverge more from the posterior, leading to higher-entropy prior distributions. While this can better cover diverse futures, it also increases the risk of sampling unrealistic samples, reflected in the slightly worse interaction metric \gls{acr}. Notably, the lowest \gls{amr} across all models is achieved by the CVAE Large Beta model (6b), trained with $\beta=0.5$. However, across all CVAE-based models, there is a consistent trend toward a slightly higher \gls{acr}, indicating that sampling from a less-constrained prior can generate less realistic combinations of the conditioning variables.

In summary, the evaluated deterministic and generative models offer complementary strengths: generative models (CVAE) achieve superior positional accuracy in terms of minSFDE and minSADE, even achieving the lowest \gls{amr} with one model. On the other hand, the deterministic models show lower \gls{acr}, indicating more robust interaction quality by avoiding unrealistic combinations.

\begin{figure*}
    \vspace{4pt}
    \centering
    \includegraphics[width=0.99\linewidth]{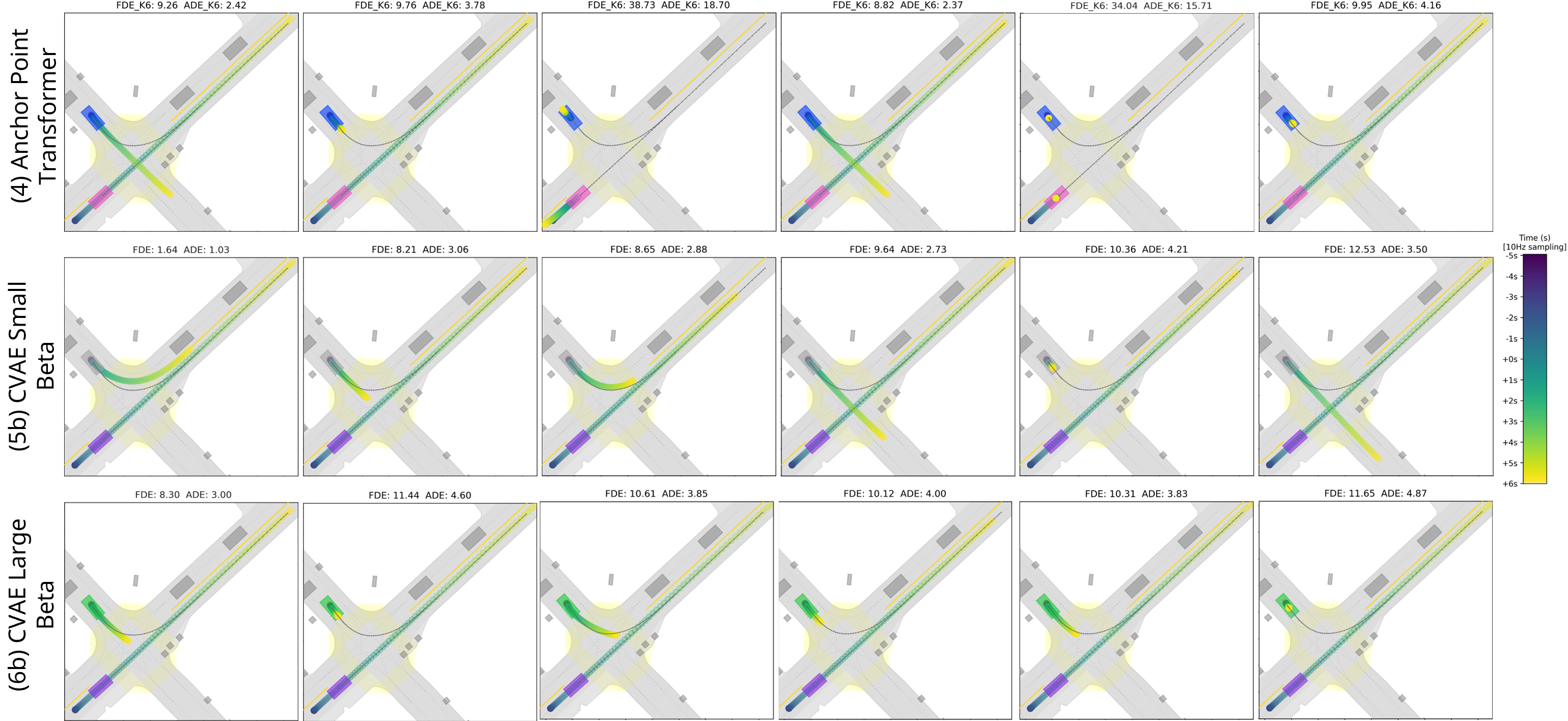}
    \caption{Scene-level multi-modal predictions from each model, along with their corresponding \gls{sfde} and \gls{sade} scores. Past and predicted future trajectories of the evaluated vehicles are shown using a color gradient, while black lines indicate the corresponding ground-truth trajectories.}
    \label{fig:trajectories}
\end{figure*}

\subsection{Multi-modality}
The metrics in Table \ref{tab5:test_metrics} assess only the mode most aligned with the ground truth, without considering the quality of the remaining $K-1$ predicted modes. To address this, we conduct an additional evaluation focusing on the diversity and multi-modality of the learned models. Specifically, we analyze the multi-modality of predictions from the Anchor Point Transformer model and the \gls{cvae} Small Beta model with mean sampling. Figure \ref{fig:trajectories} illustrates the multi-modal predictions for an example scene where one vehicle must yield to another vehicle before continuing driving.

The predictions of the deterministic Transformer-based model reveal several issues: Many predicted modes are unrealistic, such as both vehicles suddenly driving backwards, as observed in the third mode, or the vehicles stopping abruptly, as seen in the fifth mode. Additionally, even though the blue vehicle turns left in the ground truth, no predicted mode reflects this behavior, suggesting that the deterministic model struggles to capture the possible multi-modality of the scene. These issues stem from the \gls{wta} loss used during training, which optimizes only the best mode, thus ignoring the other modes. As a result, the model is not penalized for generating implausible predictions.

In contrast, the generative \gls{cvae}-based model with smaller $\beta$ produces more plausible and genuinely multi-modal predictions. The purple vehicle consistently continues straight across all modes, matching both its observed acceleration behavior and the ground truth. The second vehicle exhibits true multi-modality, displaying diverse and realistic behaviors such as yielding, turning left, or continuing straight. In every mode, it first yields to the purple vehicle before proceeding. Notably, the mode best aligned with the ground truth is the first mode, which is conditioned on the mean of the prior distribution. The consistent plausibility and diversity of the predicted modes arise from the model’s training structure. Unlike deterministic models, which predict $K$ modes in parallel while optimizing only for the best match, the \gls{cvae}-based model is trained to capture the full distribution of future outcomes, leading to more realistic and diverse predictions.

However, when compared to the predictions of the \gls{cvae}-based model with a larger $\beta$, it shows that the generative approach is sensitive to the choice of this parameter. Although with the larger $\beta$, all predicted modes remain plausible, they exhibit limited diversity. This is due to the higher weight on the KL loss, which acts as a regularizer by constraining the posterior to remain close to the prior, thereby reducing its capacity to encode useful information from the ground-truth future $\mathbf{Y}$. Consequently, the model tends to generate only minor variations of the same high-level scene, reducing the overall diversity of its predictions.

Additional prediction examples from these models are available at \href{https://frommarginaltojointpred.github.io/}{https://frommarginaltojointpred.github.io/}.

\subsection{Inference Time}
Table \ref{tab5:inference_speed} shows the measured inference times of the evaluated models, as well as the parameters of a linear regression on the number of agents $N_a$ and lane elements $N_l$ processed by the model: $T_\mathrm{inference} = \gamma_0 + \gamma_a \cdot N_a + \gamma_l \cdot N_l$. 

As expected, the baseline model with marginal mode recombination exhibits the longest inference time, nearly double that of the model with the same architecture but trained using a scene-level loss. This difference can be attributed to the additional post-processing required in the first model, which is more computationally expensive, particularly as the number of agents increases. The remaining deterministic models (2 - 4) show similar inference times, with an increase in inference time corresponding to an increase in model size. However, the number of processed agents and lane tokens only marginally influences their inference times. The generative models (5, 6) show slightly higher inference times due to the \gls{sft} layer in the deterministic decoder. As the full scene is processed per predicted mode, the deterministic decoder must process $K$ times the number of actor and lane tokens in its \gls{sft} layers, contributing to the increased inference time. Nevertheless, all models exhibit a strong inference time, enabling their application to online motion prediction.

\begin{table}[t]
    \centering
    \caption{Inference times of the proposed architectures, measured on an NVIDIA Quadro RTX 6000 \SI{24}{GB} GPU. We report means and standard deviations.}
    \resizebox{1.\linewidth}{!}{
    \begin{tabular}{cccccc}
    \toprule
    \textbf{Model} & \textbf{\begin{tabular}[c]{@{}c@{}}Inference\\ Time {[}ms{]}\end{tabular}} & \textbf{$\gamma_0$} & \textbf{$\gamma_a$} & \textbf{$\gamma_l$} & \textbf{$R^2$} \\
    \midrule
    (1)    & 31.58\textsubscript{$\pm$8.59} & $11.19$ & $6.84 \times 10^{-1}$ & $3.75 \times 10^{-2}$ & $0.8665$ \\
    (2)    & 16.31\textsubscript{$\pm$1.98} & $15.14$ & $1.02 \times 10^{-2}$ & $9.71 \times 10^{-3}$ & $0.0386$ \\
    (3)    & 17.10\textsubscript{$\pm$1.95} & $16.22$ & $7.33 \times 10^{-3}$ & $7.45 \times 10^{-3}$ & $0.0238$ \\
    (4)    & 19.51\textsubscript{$\pm$2.83} & $18.35$ & $2.28 \times 10^{-2}$ & $6.19 \times 10^{-3}$ & $0.0174$ \\
    (5, 6) & 22.66\textsubscript{$\pm$7.16} & $2.43$  & $1.82 \times 10^{-1}$ & $1.74 \times 10^{-1}$ & $0.7681$ \\
    \bottomrule
    \end{tabular}
    }
    \label{tab5:inference_speed}
\end{table}

\addtolength{\textheight}{-0.8cm}  

\section{CONCLUSIONS}

In this work, we evaluate different approaches to joint motion prediction by extending an open-source marginal prediction baseline into a joint prediction framework. Specifically, we investigate three approaches: (i) recombining marginal modes into joint predictions, (ii) incorporating a scene-level loss during training, and (iii) framing motion prediction as a generative task using a \gls{cvae}. We also examine the architectural modifications required to enable each strategy.

Experimental results show that recombining marginal modes into joint predictions already yields strong prediction performance but introduces significant computational overhead due to the required post-processing. Adapting the loss to a scene-level formulation surprisingly degrades performance. This can be mitigated with more expressive decoders, \eg the evaluated Transformer-based trajectory decoder, leading to a significant improvement in the prediction accuracy. However, they often produce implausible modes due to the \gls{wta} loss. Reframing the task as a generative problem via a \gls{cvae} adds minimal overhead while significantly improving both performance and mode plausibility. The investigated generative models achieve the best results on three of four evaluation metrics and exhibit clear multi-modality, though excessive regularization leads to mode collapse, limiting the diversity of predicted modes.

Potential future work includes applying our findings to more advanced models for further improvements, as well as exploring the adaptation of the \gls{cvae}-based approach to employ a shared latent scene variable instead of grouped actor-specific latent tokens.

\vspace{13pt}

\bibliographystyle{IEEEtran} 
\bibliography{library.bib}
\end{document}